\documentclass[default,iicol]{sn-jnl}
\usepackage{graphicx}
\usepackage{amsmath}
\usepackage{amsthm}
\usepackage{amssymb} 
\usepackage{multirow}
\usepackage{adjustbox}  
\usepackage{epsfig}
\usepackage{comment}

\newcommand{\RED}[1]{\textcolor{black}{#1}}

\jyear{2021}%

\theoremstyle{thmstyleone}%
\theoremstyle{thmstyletwo}

\theoremstyle{thmstylethree}

\begin{document}

\title[Article Title]{Controllable Dance Generation with Style-Guided Motion Diffusion}


\author[1,2]{\fnm{Hongsong} \sur{Wang}}
\author[3]{\fnm{Ying} \sur{Zhu}}
\author[1,2]{\fnm{Xin} \sur{Geng}}
\author[4,5,6,7]{\fnm{Liang} \sur{Wang}}

\affil[1]{School of Computer Science and Engineering, Southeast University, Nanjing 210096, China}
\affil[2]{Key Laboratory of New Generation Artificial Intelligence Technology and Its Interdisciplinary Applications (Southeast University), Ministry of Education, China}
\affil[3]{School of Cyber Science and Engineering, Southeast University, Nanjing 210096, China}
\affil[4]{New Laboratory of Pattern Recognition (NLPR)}
\affil[5]{State Key Laboratory of Multimodal Artificial Intelligence Systems (MAIS)}
\affil[6]{Institute of Automation, Chinese Academy of Sciences (CASIA)}
\affil[7]{School of Artificial Intelligence, University of Chinese Academy of Sciences}


\abstract{Dance plays an important role as an artistic form and expression in human culture, yet automatically generating dance sequences is a significant yet challenging endeavor. Existing approaches often neglect the critical aspect of controllability in dance generation. Additionally, they inadequately model the nuanced impact of music styles, resulting in dances that lack alignment with the expressive characteristics inherent in the conditioned music. To address this gap, we propose Style-Guided Motion Diffusion (SGMD), which integrates the Transformer-based architecture with a Style Modulation module. By incorporating music features with user-provided style prompts, the SGMD ensures that the generated dances not only match the musical content but also reflect the desired stylistic characteristics. To enable flexible control over the generated dances, we introduce a spatial-temporal masking mechanism. As controllable dance generation has not been fully studied, we construct corresponding experimental setups and benchmarks for tasks such as trajectory-based dance generation, dance in-betweening, and dance inpainting. Extensive experiments demonstrate that our approach can generate realistic and stylistically consistent dances, while also empowering users to create dances tailored to diverse artistic and practical needs. Code is available on Github: \url{https://github.com/mucunzhuzhu/DGSDP}.
}




\maketitle

\section{Introduction}
Dance is a universal form of human expression, combining rhythmic body movements with artistic and cultural elements. The ability to automatically generate realistic and stylistically consistent dance movements has significant potential applications in virtual avatars, video game design and film production. However, dance generation presents unique challenges due to the intricate spatial-temporal dynamics of human motion, the variability of dance styles, and the need for synchronization with external inputs, such as music.

Controllable generation is an essential aspect of AI-generated content, enabling users to direct and refine the creative process to meet specific requirements. In image generation, models like ControlNet~\cite{zhang2023adding} enable users to manipulate image content through various conditions and prompts. In the realm of motion generation, control mechanisms are crucial for aligning generated content with external constraints~\cite{karunratanakul2023guided,xieomnicontrol}. Without controllability, generative models risk producing human motions that lack alignment with user intent or context-specific requirements.

Dance generation, typically conditioned on music, aims to synthesize realistic and expressive dance movements based on input conditions. With the rise of deep learning and unified motion modeling~\cite{wang2025foundation,wang2025heterogeneous}, data-driven models have significantly advanced this field. Generative Adversarial Networks (GANs)~\cite{sun2020deepdance,kim2022brand} and Transformer-based architectures~\cite{li2021ai,valle2021transflower,kim2024kinematic} have been adopted to capture the sequential nature of dance movements. However, these models suffer from mode collapse, where the generator produces limited variations of motions, making them less effective for generating diverse and smooth motions.

\begin{figure}[tb]
	\centering
	\includegraphics[width=\linewidth]{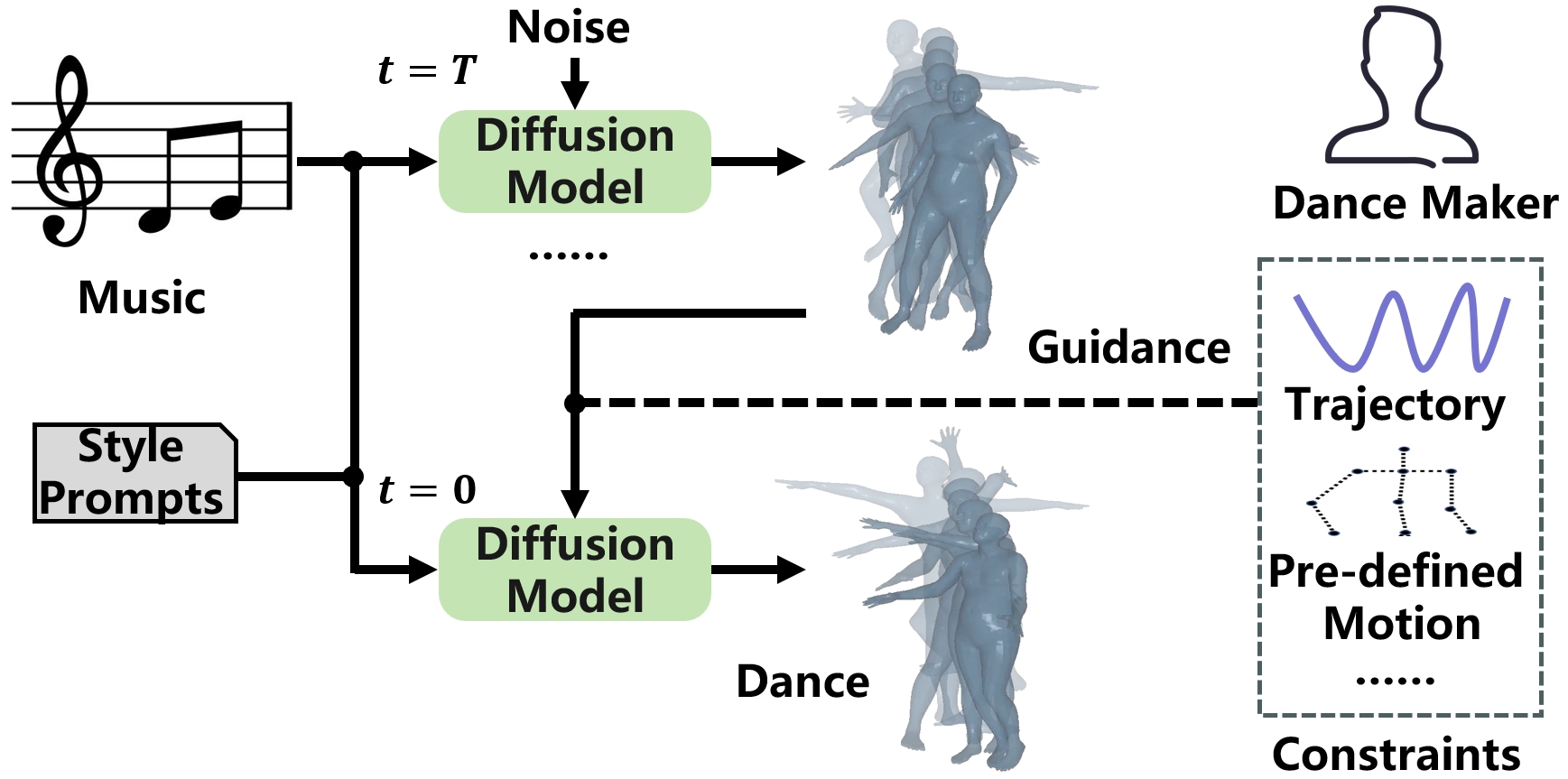}
	\caption{The illustration of controllable dance generation with style prompts and user-provided constraints. We use a diffusion-based model as an example, with controlled constraints serving as guidance for the diffusion process. }
	\label{fig:intro}
\end{figure}

Inspired by the success of diffusion-based models in human motion synthesis~\cite{tan2025sopo}, several recent works have applied diffusion models to dance generation~\cite{tseng2023edge,alexanderson2023listen,li2024lodge,Luo_2024_CVPR}. Diffusion models naturally avoid mode collapse and can generate diverse and high-quality motions. By iteratively refining motions from noise, these approaches produce smooth and realistic motion sequences based on the music input. However, they overlook the importance of style in generating expressive dances. 

Another important aspect that cannot be neglected in dance generation is the controllability. Dance, as a dynamic art form, often requires meticulous adjustments and refinements throughout its creation and performance phases. The ability to flexibly control dances can not only enhance artistic expression but also enable dancers to respond effectively to user feedback. A practical dance generation approach ought to possess sufficient flexibility to cater to a diverse range of tasks. To the best of our knowledge, there are no established benchmarks for controllable dance generation.

Since style encompasses characteristics such emotional tone, the lack of explicit style modeling limits current approaches' ability to adapt to user-defined stylistic preferences or generate expressive dance movements. Style modeling is crucial for unlocking new possibilities in controllable dance generation. To address this gap, we explicitly integrate style prompts in the diffusion-based framework, as illustrated in Fig.~\ref{fig:intro}. We also incorporate user-provided constraints as guidance to control the generated dance movements.

To this end, we introduce Style-Guided Motion Diffusion (SGMD) for controllable dance generation. The SGMD is a transformer-based diffusion framework that seamlessly integrates music conditions and style prompts into the dance generation. A lightweight Style Modulation module is introduced to enable the integration of style information without affecting the content features. A Spatial-Temporal Masking mechanism is designed to refine dance movements to adhere to specific temporal and joint constraints. To fully utilize the style information, we explore three types of style prompts: one-hot encoding, genre name, and style description prompts. The description prompt is initially generated by GPT-3 through multiple trials and manually refined to meet the requirements. To accommodate different dance generation tasks, a spatial-temporal masking is integrated in the backward diffusion process. To better align practical applications, we setup several settings and establish benchmarks under these settings.

Our contributions are as follows:
\begin{itemize}
	\item We study the important yet underexplored task of controllable dance generation and propose a style-guided motion diffusion to address it.
	\item We propose a lightweight style modulation module and a spatial-temporal masking to integrate style prompts and enforce spatial constraints, respectively, within the diffusion-based dance generation model.
	\item We set up new benchmarks for controllable dance generation tasks, such as trajectory-based dance generation, dance in-betweening, and dance inpainting, while our approach achieves state-of-the-art performance across a variety of dance generation and editing tasks.
\end{itemize}

\section{Related Work}
\noindent\textbf{Music-Conditioned Dance Generation: }
Generating dances synchronized with music in both timing and aesthetics remains a challenging task, with models such as Generative Adversarial Networks (GANs), Variational Autoencoders (VAEs), and Transformers being applied in this domain. A GAN-based framework aimed at bridging the gap between dance motion and music~\cite{sun2020deepdance}. Similarly, Li et al.~\cite{li2021ai} present a transformer-based model featuring a deep cross-modal transformer block with full-attention functionality. Valle-P{\'e}rez et al.~\cite{valle2021transflower} present a probabilistic autoregressive framework that incorporates both preceding poses and musical information. 
Huang et al.~\cite{huang2022genre} design a transformer-based system to generate genre-specific dances. Kim et al.~\cite{kim2022brand} propose a conditional GAN framework leveraging transformers to sample latent representations for diverse dance genres. 
For generating convincing long-term dance sequences, Li et al.~\cite{li2022danceformer} devise a two-phase strategy, involving the generation of key poses followed by the prediction of smooth parametric motion curves. Li et al.~\cite{siyao2022bailando} introduce an architecture with two core elements: a choreographic memory module for encoding and quantizing poses and an actor-critic GPT that sequences these into dances. Gong et al.~\cite{gong2023tm2d} explore the integration of text and music into dance generation, employing a cross-modal transformer for multimodal feature encoding. Despite these advancements, many existing approaches lack effective mechanisms for flexible editing during motion generation.


Recent works use generative diffusion \RED{or autoregressive models} to generate diverse motion sequences conditioned on audio features. 
Tseng et al.~\cite{tseng2023edge} propose a diffusion model that offers editing capabilities.
Alexanderson et al.~\cite{alexanderson2023listen} apply diffusion models to generate human motion driven by audio, leveraging Conformers for enhanced performance. 
Li et al.~\cite{li2024lodge} introduce a two-stage diffusion approach that synthesizes long dance movements in a coarse-to-fine manner. 
Luo et al.~\cite{Luo_2024_CVPR} introduce a method that improves the spatial-temporal representation of music and dance, achieving a balance between the quality and diversity of generated outputs. 
\RED{SoulNet~\cite{li2025music} integrates residual vector quantization and a cross-modal retrieval prior to produce temporally synchronized and expressive holistic dance sequences. Danceba~\cite{fan2025align} integrates phase-based rhythm extraction and temporal-gated causal attention to achieve precise beat alignment and enhance rhythmic sensitivity.} 
Guo et al.~\cite{guo2025controllable} propose a diffusion-based framework that generates 3D dance motion sequence controlled by the 2D motion sequence. Unlike the above approaches, this work concentrates on controllable dance generation by integrating style information within a diffusion framework.


\noindent\textbf{Controllable Human Motion Synthesis: } 
Compared to controllable image/video generation~\cite{zhang2023adding,wang2025uniadapter}, the field of controllable human motion synthesis is less explored. Different from human motion generation~\cite{zhu2023human,jiang2023motiongpt,meng2024rethinking} that focuses on creating realistic motion sequences, controllable human motion synthesis emphasizes generating motions that adhere to specific constraints or user-defined controls. 
Karunratanakul et al.~\cite{karunratanakul2023guided} propose Guided Motion Diffusion to incorporate spatial constraints into the motion generation process. To enable motion synthesis and editing, Kim et al.~\cite{kim2023flame} propose a diffusion-based model that allows part-based edits at both frame and joint levels without fine-tuning. OmniControl~\cite{xieomnicontrol} enables flexible, time-specific spatial control over multiple joints in text-conditioned human motion generation using diffusion models. 
MotionFix~\cite{athanasiou2024motionfix} generates motion edits as specified by the input text using a conditional diffusion model. 
DNO~\cite{karunratanakul2024optimizing} optimizes diffusion noise in pre-trained models, enabling efficient motion editing and control without task-specific training. CoMo~\cite{huang2025controllable} first decomposes motions into discrete codes and then autoregressively generates sequences of pose codes by leveraging the knowledge priors of large language models. MotionLCM~\cite{dai2025motionlcm} improves real-time controllable motion generation by enhancing runtime efficiency through one-step inference in latent diffusion models. 
\RED{A multi-task framework is developed to simultaneously perform motion editing and predict motion similarity~\cite{li2025simmotionedit}. DART~\cite{zhaodartcontrol} achieves spatial control of generated motions within a diffusion-based autoregressive framework. MotionReFit~\cite{jiang2025dynamic} enables robust spatial–temporal text-driven motion editing with an autoregressive diffusion model.}
However, these works focus on text-to-motion, and there is limited research on controllable human motion synthesis conditioned on audio inputs, which this work addresses.

\begin{figure*}[tb]
	\centering
	\includegraphics[width=1.0\linewidth]{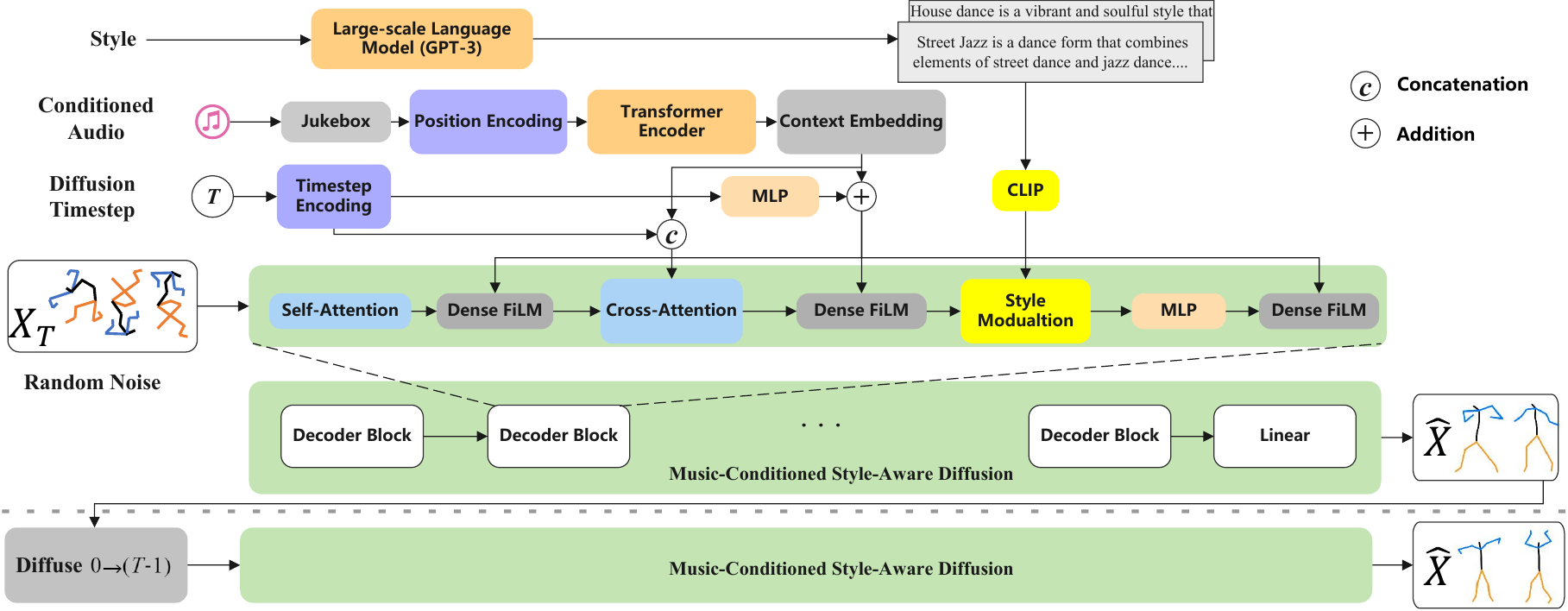}
	\caption{The proposed Style-Guided Motion Diffusion (SGMD). The model is fed with a noisy motion sequence $X_T$ of length $ N $ in a noising step $T$, along with conditioning music $c$, style prompts $s$ and $ T$ itself. The music embedding serves as input for the cross-attention module. During the inference process, SGMD samples noise $X_T$ given the conditions $c$ and $s$ to predict the clean sample $\hat{X}$. Subsequently, it diffuses this sample back to $X_{T-1}$, and repeats this iterative process until $t=0$ is reached.}
	\label{fig:model}
\end{figure*}

\section{Method}
We introduce Style-Guided Motion Diffusion (SGMD) for controllable dance generation. Enhanced by Spatial-Temporal Masking, the SGMD provides flexibility in generating and editing dance motions. The structure of SGMD is illustrated in Fig.~\ref{fig:model}.

\subsection{Preliminaries}
For human motion synthesis, auxiliary losses are frequently employed to enhance physical realism when realistic simulations are not available~\cite{tang2022real,petrovich2021action}. To encourage natural and coherent motion prediction and prevent artifacts, Tevet $et$ $al.$~\cite{tevet2022human} incorporates three geometric losses: the basic joint position loss $ {\mathcal L}_{j} $, the velocity loss $ {\mathcal L}_{v} $ and the foot contact consistency loss $ {\mathcal L}_{f} $. The definitions of these losses are as follows:
\begin{align} 
	{\mathcal L}_{j} &= \frac{1}{N} \sum\limits_{i=1}^{N} {||\mathrm {FK}(x^{(i)}) - \mathrm {FK}({\hat{x}}^{(i)})||}^2_2, \label{eq:joint} \\
	{\mathcal L}_{v} &= \frac{1}{N-1} \sum\limits_{i=1}^{N-1}{||(x^{(i+1)}-x^{(i)})-({\hat{x}}^{(i+1)}-\hat{x}^{(i)})||}^2_2, \label{eq:vel} \\
	{\mathcal L}_{f} &= \frac{1}{N-1} \sum\limits_{i=1}^{N-1}{||(\mathrm{FK}(\hat{x}^{(i+1)})-\mathrm{FK}({\hat{x}}^{(i)})) \cdot \hat{f}^{(i)}||}^2_2,  \label{eq:foot}
\end{align} 
where $\mathrm {FK}(\cdot)$ represents the forward kinematic function, which transforms joint angles into joint positions, $\hat{f}^{(i)}$ the predefined binary foot contact mask, and the superscript $(i)$ indicates the frame index. Incorporating the contact consistency loss has been shown to considerably enhance the authenticity of generated motions~\cite{tevet2022human}.

\subsection{Style-Guided Motion Diffusion}
We extend the Human Motion Diffusion~\cite{ho2020denoising} by incorporating style modulation with prompts, resulting in Style-Guided Motion Diffusion (SGMD). In addition to the music condition $c$, this model also includes the style prompts $s$. The training objective is:
\begin{equation}
	\label{eq:new_L_d}
	{\mathcal L}_{d} = {\mathbb{E}}_{x,t}[{||x-{\hat{x}}_\theta(\sigma_t,t,c,s)||}^2_2],
\end{equation}
where $\hat{x}_\theta(\cdot)$ denotes the transformer-based decoder block, which primarily comprises self-attention, feature-wise linear modulation (FiLM)~\cite{perez2018film}, cross-attention, and Style Modulation. 

The Style Modulation layer aims to increase the influence of style features on the generated dances. This module, $\mathrm{SM}(\cdot)$, enables the integration of additional style prompts into existing transformer models without affecting the content features. 
Formally, it is described as:
\begin{equation}
	\mathrm{SM}(z,s) =  \frac{z}{{||z||}_2 \cdot r} \cdot \mathrm{FC}(s), z \in {\mathbb{R}}^{T \times d_z},s \in  {\mathbb{R}}^{d_s},
\end{equation}
where $z$ refers to the input, $s$ represents the music style prompts, $r$ is a scaling factor, $\mathrm{FC}(\cdot)$ denotes the fully connected layer, $d_s$ is the embedding dimension and $d_z$ is the hidden dimension. 
Unlike adaptive instance normalization~\cite{karras2019style,karras2020analyzing}, which directly manipulates the mean and variance of input and style features to achieve style transfer, the proposed style modulation is more simple and lightweight. The normalized input preserves the content features, while the multiplication incorporates the style features. Since the style modulation module is lightweight and the prompts for a style category are typically pre-extracted, the computational cost introduced by the style prompts is negligible in the diffusion-based framework.



We train the diffusion model following the common classifier-free guidance~\cite{ho2022classifier} setting.
During training, the condition $c$ and style prompt $s$ are randomly replaced with a low probability. This training strategy enhances the alignment between the generated motions, music conditions, and style prompts. After training, the guided inference is with style prompts is:
\begin{equation}
	\widetilde{x}(\hat{\sigma}_t,t,c,s) = w \cdot \hat{x}(\hat{\sigma}_t,t,c,s) + (1-w) \cdot \hat{x}(\hat{\sigma}_t,t,\emptyset,s),
\end{equation}
where $w$ is the guidance weight with a positive value. The influence of condition $c$ can be amplified by setting $w>1$.

The overall training loss is the weighted combination defined as follows:
\begin{equation}
	{\mathcal L} = {\mathcal L}_{d} + \lambda_{j}{\mathcal L}_{j} + \lambda_{v}{\mathcal L}_{v} + \lambda_{f}{\mathcal L}_{f},
\end{equation}
where $\lambda_{j}$, $\lambda_{v}$ and $\lambda_{f}$ are weighted coefficients. The detailed training algorithm is shown in Algorithm~\ref{training_algorithm}.

\begin{algorithm}
	\caption{Training Algorithm of our Pipeline}
	\label{training_algorithm}
	\begin{algorithmic}[1]
		\Require Initialized dance generation network $\hat{x}_{\theta}$, noising steps $T$, maximum iterations $I_{max}$.
		\Require Known dance sequences $x$, music conditioning $c$, and style description prompts $s$.
		\Ensure Trained dance generation network $\hat{x}_{\theta}$.
		\For{$I = 0,1,\dots,I_{max}$}
		\State Sample $t \sim \text{Uniform}(\{1,2,\dots,T\})$.
		\State Randomly sample noise $\epsilon \sim \mathcal{N}(0, \boldsymbol{I})$.
		\State Add noise on $x$ with $\epsilon$ and parameter $\overline{\alpha}_{t}$:
		$\sigma_t = \sqrt{\overline{\alpha}_{t}}x + (1-\overline{\alpha}_{t})\epsilon$
		\State Replace $c$ with $\emptyset$ with a low probability.
		\State Calculate losses $L_j$, $L_v$, and $L_f$ in Eqs.~(\ref{eq:joint}--\ref{eq:foot}).
		\State Calculate diffusion loss $L_d$ in Eq.~(\ref{eq:new_L_d}).
		\State Update parameters of the network:
		$\theta = \theta - \nabla_{\theta}(\mathcal{L}_{d} + \lambda_{j}\mathcal{L}_{j} + \lambda_{v}\mathcal{L}_{v} + \lambda_{f}\mathcal{L}_{f})$
		\EndFor
	\end{algorithmic}
\end{algorithm}




\begin{figure}[tb]
	\centering
	\includegraphics[width=\linewidth]{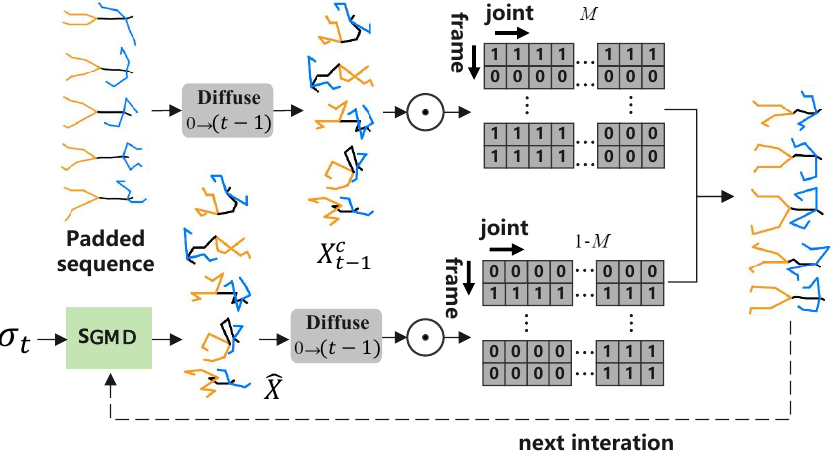}
	\caption{Controllable dance generation with spatial-temporal masking. For the known sequence, we add noise on it to obtain noisy sequence at timestep $t-1$ directly. For unknown sequences, we first use the trained network ${\hat{x}}_{\theta}$ to predict the motion at the timestep 0, and then we add noise on it to obtain noisy sequence at timestep $t-1$ . The mask is two-dimensional and allows for control in both the temporal dimension and spatial dimension.}
	\label{fig:editing_algorithom}
\end{figure}

\subsection{Spatial-Temporal Masking}
To flexibly edit generated dances, we design a spatial-temporal masking strategy, which is illustrated in Fig.~\ref{fig:editing_algorithom}. We first pad the known dance sequence to match the size of the target generation. Then, for the padded known dance sequence $x_0$, we add noise on it to obtain noisy sequence at the timestep $t-1$,
\begin{equation}
	x^{known}_{t-1} = \sqrt{\overline{\alpha}_{t-1}}x_0 + (1-\overline{\alpha}_{t-1})\epsilon,
\end{equation}
where $\epsilon \sim \mathcal N (0,\boldsymbol{I}) $. 

For an unknown sequence, the model initially predicts the target sequence from random noise, subsequently introducing noise to generate a noisy sequence at the timestep $t-1$,
\begin{equation}
	x^{unknown}_{t-1} = \sqrt{\overline{\alpha}_{t-1}}\hat{x}_{\theta}(\hat\sigma_t, t, c,s) +  (1-\overline{\alpha}_{t-1})\epsilon.
\end{equation}

Finally, $x^{known}_{t-1}$ and $x^{unknown}_{t-1}$ are combined with a spatial-temporal mask $M$,
\begin{equation}
	{\hat{\sigma}}_{t-1} = M \odot x^{known}_{t-1} + (1-M) \odot x^{unknown}_{t-1},
\end{equation}
where the mask $M \in {\{0,1\}}^{F \times J} $ can be changed arbitrarily with 0 or 1 values in both the temporal and spatial dimensions, $F$ and $J$ represent the number of frames and joints, respectively. This process is iterated by taking ${\hat{\sigma}}_{t-1}$ as the noise for the next iteration. The spatial-temporal mask supports any combination of temporal constraints and joint constraints. 
This masking strategy offers a robust tool for subsequent applications to generate dance sequences that precisely adhere to arbitrary constraints.

The spatial-temporal masking can also be applied for long-term dance generation. Since the model generates a set of frames of a dance sequence at once, increasing the maximum sequence length results in a linear increase in computational cost. In addition, dance generation requires the music condition to match the motion sequence in length, which further extends the memory requirements. We follow the strategy~\cite{tseng2023edge} to achieve temporal consistency between multiple sequences in long-term dance generation.

\subsection{Prompts of Dance Style}
Incorporating dance style is essential for producing high-quality, emotional and contextually appropriate dance movements. For the Style-Guided Motion Diffusion, we explore three encoding methods for prompts of dance style.

\noindent \textbf{One-Hot Encoding:} One-hot encoding is commonly used to handle categorical variables, particularly when the number of categories is limited. As the music style can be divided into categories, we use one-hot encoding to represent the style information.

\noindent \textbf{Genre Name:} To obtain a more semantic representation, we utilize CLIP~\cite{radford2021learning} to extract features for the words associated with the music genre. This embedding captures deep contextual relationships within the textual genre, resulting in more accurate feature representations.

\noindent \textbf{Style Description:} Large-scale language models pre-trained on massive text~\cite{liu2023pre} demonstrate impressive capabilities for text generation tasks~\cite {yoo2021gpt3mix}. To get a more detailed characteristics of the music style, we use GPT-3~\cite{brown2020language} to generate style description for dance. We choose the following function to acquire the style description prompts: ``Please generate a detailed description of the dance [$g$], including the characteristics of the dance in terms of body movement.", where $g$ is the dance genre.

\section{Experiments}
\subsection{Dataset and Evaluation Metrics}
The AIST++ dataset~\cite{li2021ai} includes 1,408 high-quality dance movements synchronized with music from 10 genres. It comprises motion sequences of varying durations, ranging from 7 seconds to 50 seconds, with an average duration of 13 seconds. We follow the train/test splits of EDGE~\cite{tseng2023edge}. Instances in the training set are cut to 5 seconds at 30 FPS with a stride of 0.5 seconds, and instances in the testing set are cut to 5 seconds at 30 FPS with a stride of 2.5 seconds.

The Beat Alignment Score ($BeatAlign$), Physical Foot Contact Score ($PFC$), Frechet Inception Distance ($FID$) and Diversity ($Div$) are used as evaluation metrics. 

\noindent \textbf{Beat Alignment Score:} It is used to measure the alignment between the music and the generated movement. We calculate the average time distance between each music beat and its closest dance beat as the beat alignment score. 

\noindent \textbf{Physical Foot Contact Score:} It is a quantitative acceleration-based metric used to score the physical plausibility of the generated kinematic motion. 

\noindent \textbf{Frechet Inception Distance(FID):} FID is the distance between the generated dance movement distribution and the real movement distribution. We calculate the FIDs between the generated dances and the AIST++ dataset for all motion sequences in the kinetic feature space ($FID_k$) and geometric feature space ($FID_g$).

\noindent \textbf{Diversity:} This metric assesses the diversity of dance motions by computing the average Euclidean distance between different generated dance motions. Similarly, we compute the Diversity between all motion sequences of the generated dances in the kinetic feature space ($Div_k$) and the geometric feature space ($Div_g$).

\subsection{Experimental Setup}

\noindent \textbf{Implementation Details: }The proposed model has 49.7 million (M) parameters, and was trained on one NVIDIA RTX A6000 GPU for 2 days with a batch size of 128. The implementation is based on the PyTorch. The number of iterations is 1000. The learning rate is 0.0002 and the weight decay is 0.02. The music condition transformer encoder has a depth of 2 layers, with 8 attention heads and a dimension of 512. The decoder block has a depth of 8 layers, with 8 attention heads and a dimension of 512. For long-term generation, we use linear weighted summation to sum up corresponding parts of different slices (5s). We perform a weighted summation of the latter half (2.5s) of the previous slice and the former half (2.5s) of the next slice. 
To mitigate the impact of randomness of stochastic, all evaluation metrics obtained by averaging over 100 trials. 

\begin{table*}[htbp]
	\centering
	\caption{Comparision of dance generation on the AIST++ dataset. For simplicity, $\uparrow$ means higher is better, $\downarrow$ means lower is better and $\rightarrow$ means closer to the ground truth is better. During inference, akin to EDGE~\cite{tseng2023edge}, the guidance weight $w$ is set to 1 by default, and the results obtained with $w = 2$ are also reported to explore the influence of amplified conditions. }
	\label{tab:Resluts}
	\adjustbox{max width=\textwidth}{ 		\begin{tabular}{l|c|cccccc|cccccc}
			\hline
			\multirow{2}[0]{*}{Method} & \multirow{2}[0]{*}{Model} & \multicolumn{6}{c|}{Guidance weight $w$ = 2}  & \multicolumn{6}{c}{Guidance weight $w$ = 1} \\
			\cline{3-14}
			& & $BeatAlign\uparrow $  & $PFC\downarrow$   & $FID_k\downarrow$ & $FID_g\downarrow$ & $Div_k\rightarrow$& $Div_g\rightarrow$ & $BeatAlign\uparrow $  & $PFC\downarrow$   & $FID_k\downarrow$ & $FID_g\downarrow$ & $Div_k\rightarrow$& $Div_g\rightarrow$ \\
			\hline
			\RED{Ground Truth} & -- & 0.35  & 1.33  & --     &  --    & 9.43  & 7.33 & 0.35  & 1.33  & --     &  --    & 9.43  & 7.33  \\
			\hline
			FACT~\cite{li2021ai} & Others & -- & -- & -- & -- & -- & --  & 0.20 & 30.39 & 561.14 & 170.36 \\ 
			Bailando~\cite{siyao2022bailando} & VQ-VAE & -- & -- & -- & -- & -- & --  & 0.21  & 1.72  & 24.30  & 20.81  & 6.83  & 7.69  \\
			TM2D~\cite{gong2023tm2d} & VQ-VAE  & -- & -- & -- & -- & -- & -- & 0.19 & 3.28 & \textbf{15.37} & 28.35 & \textbf{9.10} & \textbf{7.91} \\
			\hline
			EDGE~\cite{tseng2023edge} & Diffusion & 0.26  & 1.56  & 35.35 & \textbf{18.92} & 5.32  & 4.90 & 0.25  & \textbf{1.19} & 45.41 & 19.42 & 4.82  & 5.24 \\
			Lodge~\cite{li2024lodge} & Diffusion & -- & -- & -- & -- & -- & --   & 0.24  & -- & 37.09  & 18.79  & 5.58  & 4.85 \\
			\textbf{SGMD (ours)} & Diffusion & \textbf{0.31} &\textbf{1.51}& \textbf{34.32} & 19.04 & \textbf{6.29} & \textbf{5.81} & \textbf{0.28} & \textbf{1.65}  & \textbf{37.21} & \textbf{18.19} & \textbf{7.25} & 6.42 \\
			\hline
	\end{tabular}}
\end{table*}

\begin{table*}[htbp]
	\centering
	\caption{Results of long-term dance generation on the AIST++ dataset. We note that very few works focus on long dance generation; however, their settings and corresponding models are not released. Therefore, we do not include comparisons with those methods.}
	\label{tab:Long-term results}
	\adjustbox{max width=\textwidth}{ 
		\begin{tabular}{l|cccccc|cccccc}
			\hline
			\multirow{2}[0]{*}{Method} & \multicolumn{6}{c|}{7.5 seconds}  & \multicolumn{6}{c}{10 seconds} \\
			\cline{2-13}
			& $BeatAlign\uparrow $  & $PFC\downarrow$   & $FID_k\downarrow$ & $FID_g\downarrow$ & $Div_k\rightarrow$& $Div_g\rightarrow$ & $BeatAlign\uparrow $  & $PFC\downarrow$   & $FID_k\downarrow$ & $FID_g\downarrow$ & $Div_k\rightarrow$& $Div_g\rightarrow$ \\
			\hline
			\RED{Ground Truth} & 0.38  & 1.04  & --     & -- & 9.34 & 7.47 & 0.49  & 1.70  & --  & --  & 9.34 & 7.47 \\
			\hline
			EDGE~\cite{tseng2023edge} & 0.26     &    \textbf{1.11}   &   59.43  &   25.60 &    2.97   & 3.52 & 0.25  & \textbf{0.88} & 68.63 & 32.24 & 2.55 & 3.14 \\  
			\textbf{SGMD (ours)} & \textbf{0.30}    &  1.38    &  \textbf{55.65}    &    \textbf{20.74}   &  \textbf{4.10}   & \textbf{5.74} & \textbf{0.30} & 2.13 & \textbf{56.15} & \textbf{29.98} & \textbf{5.21} & \textbf{6.45} \\
			\hline
	\end{tabular}}
\end{table*}

\noindent \textbf{Settings of Controllable Dance Generation: }As there are no relevant benchmarks, we set up experiments for controllable dance generation by integrating spatial constraints. The settings for controllable generation are as follows.



\noindent \textbf{Trajectory:} Given a predefined trajectory, the model generates dance movements that are coherent with the trajectory.

\noindent \textbf{Seed Motion:} Given motions of the first and second frames, the model generates motions of the following frames.

\noindent \textbf{In-betweening:} Given motions of the first and last frames, the model completes motions of the in-between frames.

\noindent \textbf{Inpainting:} Given a random mask, the model completes the remaining frames based on the mask. We control the model to randomly give 70\% of the true sequences.

\noindent \textbf{Upper-body Generation:} Since the mask possesses both spatial and temporal dimensions, apart from controlling it in the temporal dimension, we can also manipulate it in the spatial dimension. Upper-body generation aims to generate motions of the upper-body given lower-body motions.

\noindent \textbf{Lower-body Generation:} Given upper-body dance movements of all sequences, the model automatically generates lower-body dance movements.

\begin{table*}[htbp]
	\centering
	\caption{Results of controllable dance generation given different spatial constraints.}
	\adjustbox{max width=1\textwidth}{ 
		\begin{tabular}{l|l|cccccc}
			\hline
			\multicolumn{1}{l|}{Conditions} & Method & $BeatAlign\uparrow $  & $PFC\downarrow$   & $FID_k\downarrow$ & $FID_g\downarrow$ & $Div_k\rightarrow$& $Div_g\rightarrow$\\
			\hline
			-- & \RED{Ground Truth} & 0.35  & 1.33  & --     &  --    & 9.43  & 7.33 \\
			\hline		
			\multicolumn{1}{l|}{\multirow{2}[2]{*}{Trajectory}} 
			& EDGE~\cite{tseng2023edge}  & 0.27& 1.51& \textbf{31.1} & 16.23& 5.03& 5.23 \\
			& \textbf{SGMD (ours)} &  \textbf{0.33} & \textbf{1.23} & 32.22& \textbf{14.85} & \textbf{5.19} & \textbf{6.02}  \\
			\hline
			\multicolumn{1}{l|}{\multirow{2}[2]{*}{Seed motion}} 
			& EDGE~\cite{tseng2023edge}  &   0.27    &    1.19   &   32.62    &   21.68    &    5.81   &  5.15 \\
			& \textbf{SGMD (ours)} &   \textbf{0.33}    &     \textbf{1.18}   &  \textbf{27.05}     &   \textbf{17.36}     &   \textbf{7.34}     & \textbf{5.79}  \\
			\hline
			\multicolumn{1}{l|}{\multirow{2}[2]{*}{In-betweening}} 
			& EDGE~\cite{tseng2023edge}  &  0.28      &   \textbf{1.17}    &  37.33   &  15.16     &  4.58     & 4.86 \\
			& \textbf{SGMD (ours)} & \textbf{0.34}      &   1.24    &   \textbf{30.20}    &      \textbf{14.75} &   \textbf{5.80}    & \textbf{6.16}  \\
			\hline
			\multicolumn{1}{l|}{\multirow{2}[2]{*}{Inpainting}} 
			& EDGE~\cite{tseng2023edge}  &   0.36    &  2.15     &    \textbf{21.57}  &   \textbf{17.50}     &    \textbf{7.47}   & 8.36 \\
			& \textbf{SGMD (ours)} &  \textbf{0.37}      &    \textbf{1.71}   &   23.80   &    18.78  & 7.03      &  \textbf{8.34} \\
			\hline
			\multicolumn{1}{l|}{\multirow{2}[2]{*}{Upper-body generation}} 
			& EDGE~\cite{tseng2023edge}  &  0.29     &    1.67   &   32.90  &    14.82   &    5.05   &   5.45 \\
			& \textbf{SGMD (ours)} &   \textbf{0.33}    &  \textbf{1.40}     &  \textbf{32.86}     &    \textbf{14.36}   &  \textbf{5.39}     & \textbf{6.11} \\
			\hline
			\multicolumn{1}{l|}{\multirow{2}[2]{*}{Lower-body generation}} 
			& EDGE~\cite{tseng2023edge} & 0.33  & \textbf{0.92} & \textbf{39.81} & 15.44 & 5.28 & 6.86 \\
			& \textbf{SGMD (ours)} & \textbf{0.36} & 0.99 & 42.26 & \textbf{14.73} & \textbf{5.39} & \textbf{7.37} \\
			\hline
	\end{tabular}
    }
	\label{tab:Edit_results}
\end{table*}

\subsection{Evaluation of Dance Generation}
\noindent\textbf{Results of Dance Generation: }We choose EDGE~\cite{tseng2023edge} as our baseline. Since the reported results are averaged over 100 trials to reduce randomness, we re-implement the results of EDGE and FACT on the AIST++ dataset. For different approaches, we generate 20 pieces of dances with 5-second length, covering 10 different genres. Quantitative results are summarized in Table~\ref{tab:Resluts}.
When comparing our approach with state-of-the-art methods, we observe a significant improvement in the Beat Alignment Score, the primary metric for dance generation, suggesting that the dances generated of our method are more in tune with the musical melody. Specifically, our method performs favorably against EDGE on most metrics. 
Our method achieves the lowest $FID_k$ among diffusion-based models. This metric, defined based on motion velocities and energies, reflects the physical characteristics of dance. The superior $FID_k$ score demonstrates that the proposed SGMD generates more realistic dance movements.
Our approach also achieves the closest approximation to the ground-truth values of $Div_k$ and $Div_g$ among diffusion-based models, demonstrating its ability to generate more diverse dance movements instead of converging to a limited set of fixed future motions.

\noindent\textbf{Results of Long-Term Dance Generation: }
Since both the training and test sets are composed of five-second slices, in order to validate the ability of our model to generate long-term dance sequences, we generate the corresponding dances for music that are 7.5 seconds and 10 seconds. We select the pieces of dances that are longer than 7.5 seconds or 10 seconds for long-term dance generation. Experimental result of long-term dance generation on AIST++ is shown in Table~\ref{tab:Long-term results}. 
Since there is limited work on long dance generation, we compare our approach only with the baseline.

Our method performs significantly better than EDGE~\cite{tseng2023edge} on most metrics. Specifically, our model achieves a higher Beat Alignment Score, maintaining good performance on the correlation between dance movements and music. Our model also performs better than EDGE on metrics of $FID_k$, $FID_g$, $Div_k$ and $Div_g$. This reveals that SGMD can also generate long dance movements with better quality and better diversity. It is worth noting that the performance of our model on $Div_k$ and $Div_g$ do not decrease significantly over time like EDGE. This suggests that long dance movements generated using our method can ensure diversity, avoiding repetitive movements.

\noindent\textbf{Results of Controllable Dance Generation: }
Our approach is applicable to controllable dance generation given different spatial constraints. We quantitatively compare our approach with EDGE~\cite{tseng2023edge} in Table~\ref{tab:Edit_results}. 

Given the predefined trajectory, our method outperforms EDGE in terms of the Beat Alignment Score by 0.06, representing a significant relative improvement of 22\%. Given conditions of both the seed motion and the upper-body, our method performs favorably against EDGE~\cite{tseng2023edge} on all metrics. For dance in-betweening, dance inpainting and lower-body generation, our method performs better than EDGE~\cite{tseng2023edge} on most metrics. This suggests that our approach has a strong capacity for flexible generation from both the temporal and spatial perspectives. Furthermore, it is found that the performance of our model on Beat Alignment Score is close to the ground truth or even exceeds it, revealing the potential for self-improvement properties in harmonising music and movement. 
Overall, our approach is suitable for a wide range of tasks of controllable dance generation.

\begin{table}[tbp]
	\centering
	\caption{Comparison of results with different style prompts. }
	\adjustbox{max width=0.48\textwidth}{ 
		\begin{tabular}{l|cccc}
			\hline
			Style Prompts & \multicolumn{1}{c}{$BeatAlign \uparrow $} & \multicolumn{1}{c}{$PFC\downarrow$} & \multicolumn{1}{c}{$FID_k\downarrow$} & \multicolumn{1}{c}{$FID_g\downarrow$}  \\
			\hline
			One-hot Encoding &  0.29     &     1.61  &   40.13    &  19.18 \\
			Genre Name& 0.26  & 1.96  & 32.36  & \textbf{17.54}   \\
			Style Description & \textbf{0.31}  & 1.51  & 34.32  & 19.04  \\
			\hline
	\end{tabular}
   }
	\label{tab:style_embedding}
\end{table}



\begin{table}[tbp]
	\centering
	\caption{The impact of different audio representations. }
	\adjustbox{max width=0.5\textwidth}{ 
		\begin{tabular}{l|cccccc}
			\hline
			\multirow{2}[0]{*}{Method} & \multicolumn{1}{c}{ $ BeatAlign $} & \multicolumn{1}{c}{ $PFC$} & \multicolumn{1}{c}{$FID_k$} & \multicolumn{1}{c}{$FID_g$} & \multicolumn{1}{c}{$Div_k$} & \multicolumn{1}{c}{$Div_g$} \\
			&  $\uparrow$     &  $\downarrow$     &  $\downarrow$     &   $\downarrow$    &     $\rightarrow$  & $\rightarrow$  \\
			\hline
			Jukebox & \textbf{0.31} & 1.51  & 34.32 & 19.04 & 6.29  & 5.81 \\
			Encodec & 0.28  & \textbf{1.36}  & 43.67 & 18.97 & 4.41  & 5.08 \\
			Librosa & 0.29  & 1.48  & 32.78 & \textbf{16.60} & 5.80  & \textbf{6.54} \\
			\hline
	\end{tabular}
    }
	\label{tab:audio_rep}
\end{table}

\begin{table}[t]
	\centering
	\caption{Ablation studies of the proposed SGMD. For simplicity, style modulation and style description prompt are shortened to modulation and description, respectively. } 
	\adjustbox{max width=0.5\textwidth}{ 
		\begin{tabular}{l|cccccc}
			\hline
			\multirow{2}[0]{*}{Method} & \multicolumn{1}{c}{ $ BeatAlign $} & \multicolumn{1}{c}{ $PFC$} & \multicolumn{1}{c}{$FID_k$} & \multicolumn{1}{c}{$FID_g$} & \multicolumn{1}{c}{$Div_k$} & \multicolumn{1}{c}{$Div_g$} \\
			&  $\uparrow$     &  $\downarrow$     &  $\downarrow$     &   $\downarrow$    &     $\rightarrow$  & $\rightarrow$  \\
			\hline
			Baseline & 0.26  & 1.56  & 35.35 & 18.92 & 5.32  & 4.90 \\
			w/ Modulation  & 0.26  & 1.96  & 32.36 & 17.54 & \textbf{6.46} & 5.22 \\
			w/ Description & 0.29  & 1.71  & \textbf{31.57} & 18.47 & 5.84  & 5.30 \\
			Ours & \textbf{0.31} &\textbf{1.51}& \textbf{34.32} & 19.04 & \textbf{6.29} & 
			\textbf{5.81} \\
			\hline
	\end{tabular}
    }
	\label{tab:ablation}
\end{table}

\subsection{Ablation Study and Analysis}
We conduct detailed ablation studies and analyze the impacts of various audio representations and style prompts.

\noindent{\bf Comparisons of Different Style Prompts: } In Table~\ref{tab:style_embedding}, we compare three types of style prompts: one-hot encoding, genre name and style description. We find that using style description prompts performs better than the other two prompts on most metrics. Compared to one-hot encoding and genre name, style description prompts contain richer semantic information related to musical styles, which enhances the quality of dance generation. 
Since the textual style description is only used for Style Modulation, which is lightweight, the computational cost introduced by this module can be ignored.

\noindent{\bf Comparisons of Different Audio Representation:}
In Table~\ref{tab:audio_rep}, we evaluate the impact of three independent music feature extractors: Jukebox~\cite{dhariwal2020jukebox}, Encodec~\cite{defossez2022high}, and Librosa~\cite{mcfee2015librosa}.
Jukebox, a pre-trained generative model designed for music generation, showcasing notable performance in tasks specific to music prediction as evidenced in prior studies~\cite{tseng2023edge}. Encodec is a pre-trained model, which has been used as music conditional feature extractor in dance generation tasks as well.  
Librosa is a library for analyzing audio signals.
We observe that the Beat Alignment Score and $Div_k$ achieve the optimal performance when Jukebox is utilized as the feature extractor; PFC attains its peak performance with Encodec serving as the feature extractor; and $FID_k$, $FID_g$ and $Div_g$ exhibit the best results when Librosa is employed as the feature extractor. Since the Beat Alignment Score is the primary metric, we use Jukebox to extract audio features.

\noindent{\bf Effectiveness of Style Modulation and Description Prompts: } Ablation studies of the proposed SGMD are shown in Table~\ref{tab:ablation}. To verify the effectiveness of the Style Modulation module, we remove the Style Modulation module and simply replace it with concatenating music conditions $c$ and style description prompts $s$, which then enters the Transformer-based block. We find that this variant, referred to as w/ modulation, does not improve the Beat Alignment Score over the baseline. 

In addition, we use genres instead of style description prompts, and feed its encoding to the style modulation. This approach, denotes as w/ description, also performs worse on most metrics compared to ours. The possible reason for this is that genres contain less information than style description prompts. To investigate the importance of combining Style Modulation and Style Description Prompts, we remove the Style Modulation module and omit the use of any prompts, which corresponds to the baseline. We can see that the performances are obviously worse in this case. We can conclude that it is quite important for Style Modulation and Style Description Prompts to work together.

\noindent{\bf Limitations of Evaluation Metrics: } 
Developing automated metrics for the quality of the dance generated is a challenging endeavour due to the complex, subjective and even culturally specific context of dance practice. 
FID-based metrics are suitable for data-rich domains such as image generation, but the AIST++ test set is too small to cover the train distribution. Moreover, because of the limited availability of data, both $FID_k$ and $FID_g$ rely on heuristic feature extractors that merely calculate surface-level characteristics of the data. We argue that the idea of assessing the difference between two distributions of features of dance movements is not necessarily fundamentally flawed, but that more representative features may lead to reliable automated quality assessments.

The beat alignment score calculates the average time distance between each music beat and its closest dance beat. However, dancing is not solely about syncing local speed minima in joints with beats. Instead, musical beats serve as a flexible reference for timing, rhythm, and the smooth transition between dance movements and steps. While this metric has driven progress on this issue in the past, more dance-specific metrics are needed to realize breakthroughs.

\begin{figure}[t]
	\centering
	\includegraphics[width=1\linewidth]{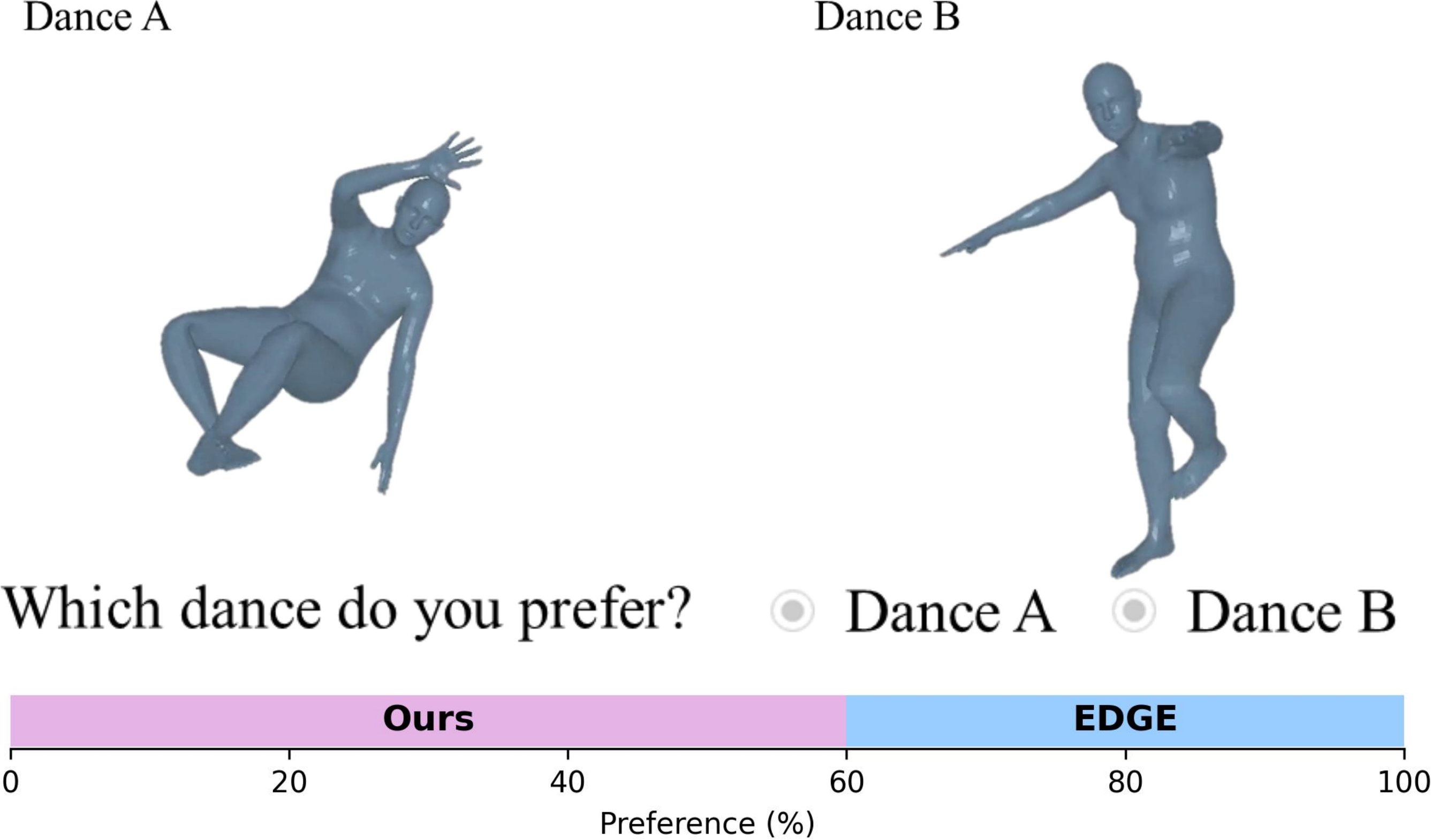}
	\caption{\RED{User study interface of dance generation and the corresponding results.}}
	\label{fig:user_study}
\end{figure}

\begin{figure}[t]
	\centering
	\includegraphics[width=1\linewidth]{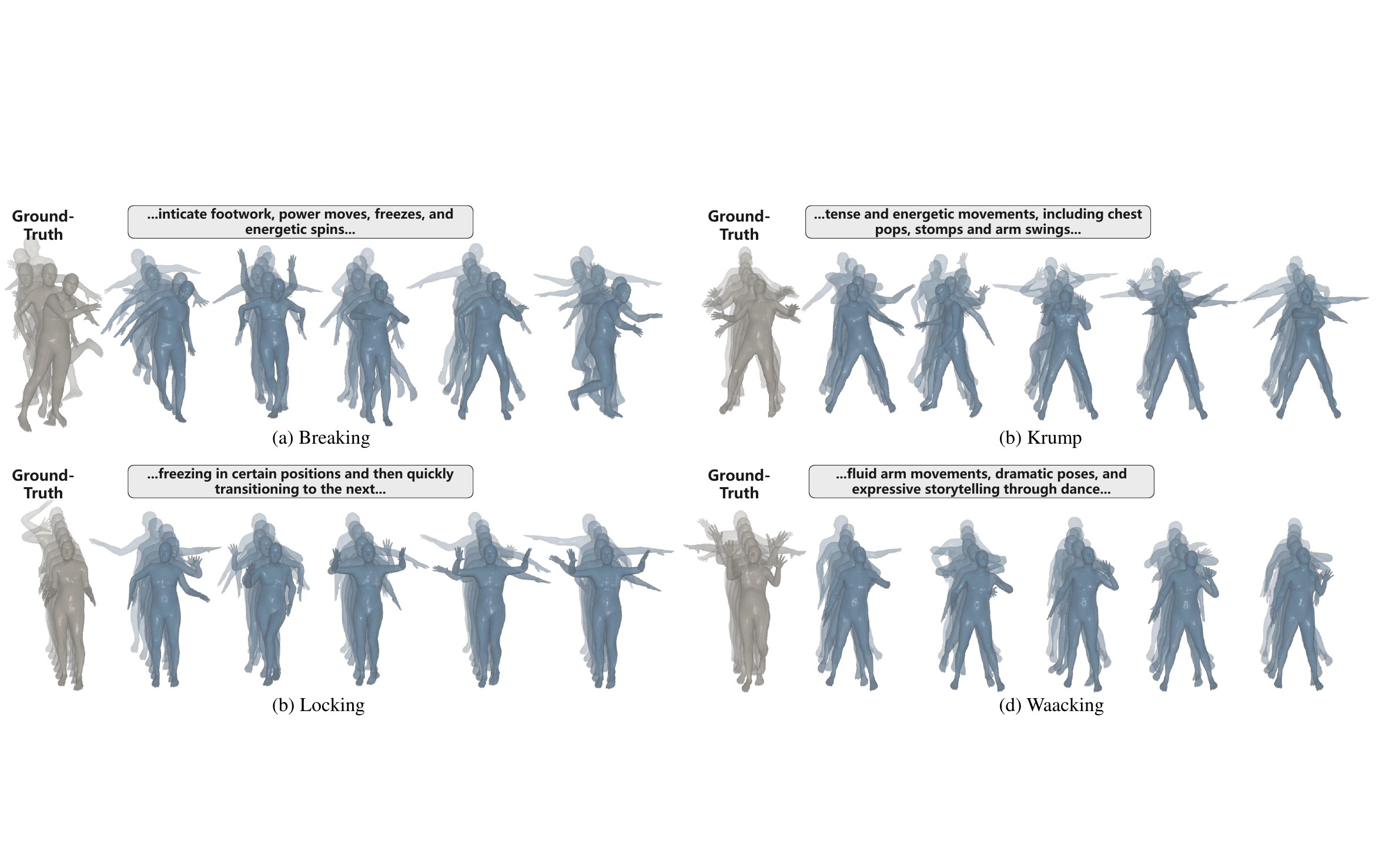}
	\caption{Visualization of generated dances for the same piece of music. We list four music genres, each associates with five different dance movements. Generated dance movements are in blue and real movements are in grey.
	}
	\label{fig:visualization}
\end{figure}

\begin{figure}[t]
	\centering
	\includegraphics[width=1\linewidth]{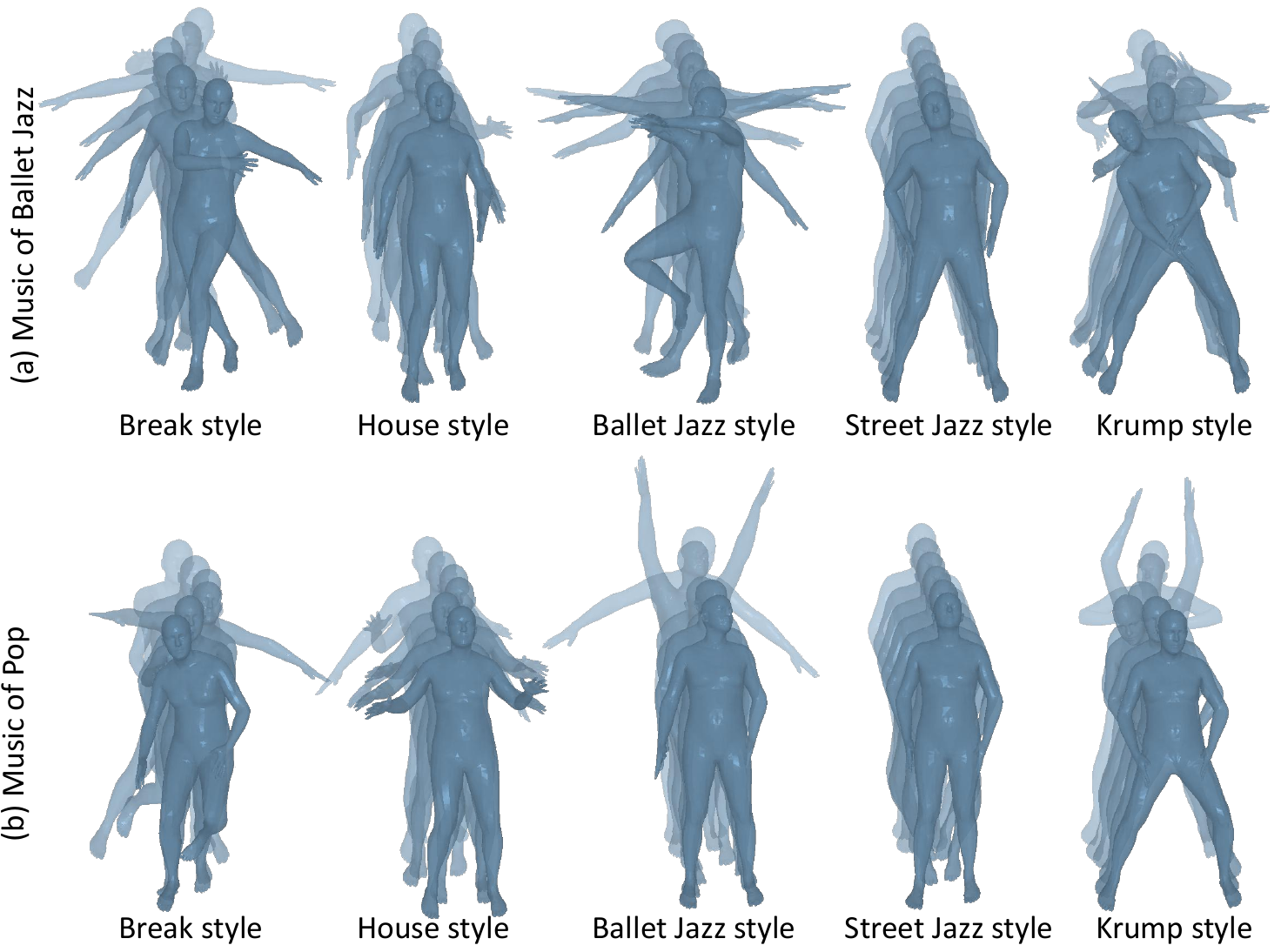}
	\caption{Visualization of generated dances for the same piece of music with different style prompts. In each example, we present dances in five different styles.
	}
	\label{fig:multi-style2}
\end{figure}

\noindent{\bf {User Study:} } 
In addition to the quantitative experiments, \RED{we invite 16 human users} to assess the visual quality of dance movements created by our method and by EDGE~\cite{tseng2023edge}. These users come from a diverse range of regions, ages, and genders. \RED{We present 21 pairs of comparison videos with the ground truth data from the AIST++ test set.} When presented with dance movements generated from the same music clip, the users are asked: ``Which dance do you prefer?" 
\RED{The user study interface and the corresponding results are shown in Fig.~\ref{fig:user_study}.
}
Our findings reveal that 60\% of the users prefer dance movements generated by our method compared to those of EDGE. Additionally, the movements created by our method exhibit greater diversity, and even some samples surpass the ground truth in quality. 

\noindent{\bf Visualizations: } For the same music, our approach can generate a variety of sensible dance movements. Some of the dance visualization results are shown in Fig.~\ref{fig:visualization}. Here, we list generated dances for two types of music: Breaking and Locking. For each type, we list five dance movements. As can be seen from the figure, our approach can generate diverse and plausible dance movements.

We also provide qualitative results of generated dances of arbitrary styles Fig.~\ref{fig:multi-style2}. Our method is capable of generating more diverse and expressive dances, holding considerable potential for creative dance generation and editing.


\subsection{Limitations and Applications}
\noindent{\bf Limitations:} \RED{One limitation of our approach is that the textual description may lack explicit rhythm-aware information, causing the text module to misalign with the rhythmic beats present in the music. Another limitation is that our model is constrained to a diffusion-based dance generation pipeline, and extending it to an autoregressive framework would be a promising direction for future work.
}

\noindent{\bf Applications:} 
This work is applicable to many interesting and promising applications, some of which are outlined below:
 
The proposed method can be applied to \textbf{interactive dance generation} by allowing users to adjust style, timing, and movement constraints, enabling personalized and adaptive dance performances to meet artistic and practical needs.

Our method has the potential for \textbf{creative dance generation}, which emphasizes the production of movements with artistic style, rhythm, and emotional expression. This task typically involves modeling complex spatial-temporal patterns and promoting motion diversity beyond mere physical realism.


\section{Conclusion}

This work studies controllable dance generation, an important yet often neglected task. We introduce Style-Guided Motion Diffusion (SGMD), which enhances the existing diffusion-based dance generation framework with a Style Modulation module and Spatial-Temporal Masking. We establish new experimental settings for controllable dance generation under different spatial constraints. Extensive experiments demonstrate that our approach achieves state-of-the-art performance across a variety of tasks, including long-term dance generation, trajectory-based dance generation, dance in-betweening, and dance inpainting. We also explore different style prompts, including one-hot encoding, genre name, and style description, and experimentally find that style description, with rich semantic information, yield the best performance. Ablated experiments demonstrate the effectiveness of the modulation module and style prompts. We also compare results of dance generation using different audio feature extractors. To obtain dances that align better with the music, we suggest using Jukebox to extract audio representations and applying style description prompts augmented by large language models. The proposed framework is suitable for flexible applications of controllable dance generation, enabling the user to freely edit or generate the desired dance sequence. We hope that this work will inspire further research in the fields of automatic and interactive dance generation.

\section*{Acknowledgments}
This research was supported by the Jiangsu Science Foundation (BK20230833, BK20243012, BG2024036), the National Science Foundation of China (62302093, 52441503, 62125602, U24A20324, 92464301), the Fundamental Research Funds for the Central Universities (2242025K30024) and the Open Research Fund of the State Key Laboratory of Multimodal Artificial Intelligence Systems (E5SP060116). 


\bibliographystyle{sn-mathphys}
\bibliography{IEEEbib}



\begin{figure}[h]
	\centering
	\includegraphics[width=0.2\textwidth]{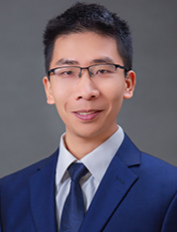}
\end{figure}

\noindent{\bf Hongsong Wang} received Ph.D. degree in Pattern Recognition and Intelligent Systems from Institute of Automation, University of Chinese Academy of Sciences in 2018. He was a postdoctoral fellow at National University of Singapore in 2019. He was a research associate at Inception Institute of Artificial Intelligence, Abu Dhabi, UAE in 2020. He is an Associate Professor with Department of Computer Science and Engineering, Southeast University, Nanjing, China.

His research interests include human action understanding, human motion modeling, motion genneration, and etc.

E-mail: hongsongwang@seu.edu.cn 

ORCID iD: 0000-0002-9464-1778

\begin{figure}[h]
	\centering
	\includegraphics[width=0.2\textwidth]{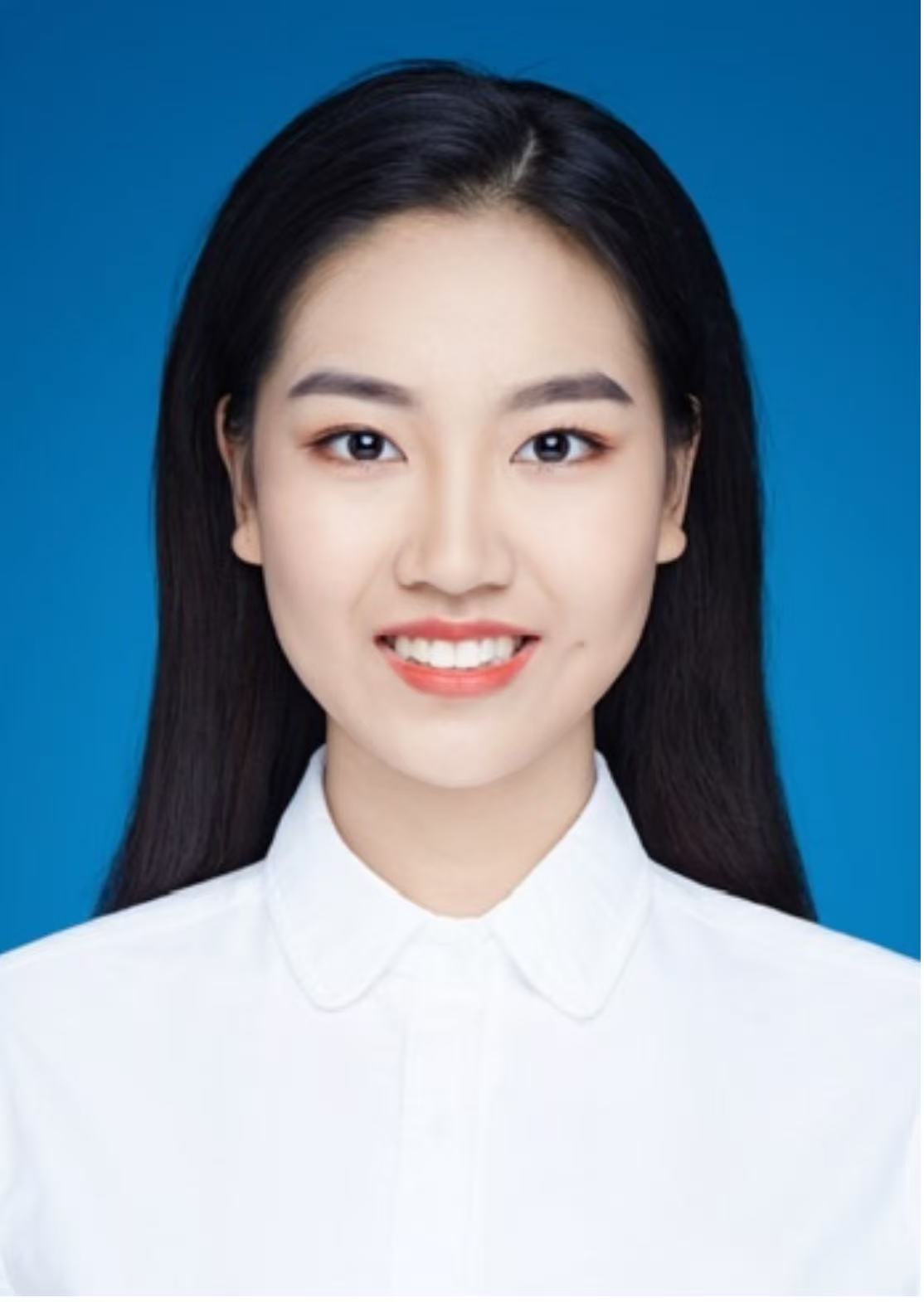}
\end{figure}

\noindent{\bf Yin Zhu } received the B.Sc. in software engineering from Nanjing University of Posts and Telecommunications, China. She is a master's degree candidate at the School of Cyber Science and Engineering, Southeast University, China.

Her research interests include human action understanding and motion generation.

E-mail: zhuy@seu.edu.cn

\begin{figure}[h]
	\centering
	\includegraphics[width=0.2\textwidth]{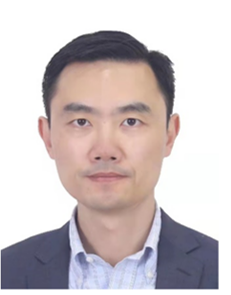}
\end{figure}

\noindent{\bf Xin Geng} has been an Associate Editor or a Guest Editor of several international journals, such as FCS, PRL, and IJPRAI. He has served as a Program Committee Chair of several international/national conferences and a Program Committee Member for a number of top international conferences, such as IJCAI, NIPS, CVPR, ICCV, AAAI, ACMMM, and ECCV.

His research interests include pattern recognition, machine learning, and computer vision.

E-mail: xgeng@seu.edu.cn 

ORCID iD: 0000-0001-7729-0622

\begin{figure}[h]
	\centering
	\includegraphics[width=0.2\textwidth]{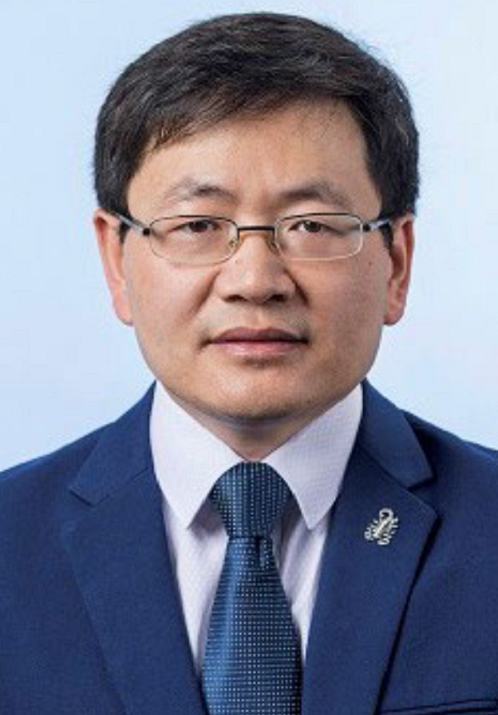}
\end{figure}

\noindent{\bf Liang Wang} received the Ph.D. degree from the Institute of Automation, Chinese Academy of Sciences (CASIA), in 2004. He is currently a Full Professor with the National Laboratory of Pattern Recognition, CASIA. He has widely published in highly ranked international journals, such as IEEE TRANSACTIONS ON PATTERN ANALYSIS AND MACHINE INTELLIGENCE and IEEE TRANSACTIONS ON IMAGE PROCESSING, and leading international conferences, such as CVPR, ICCV, and ICDM.

His research interests include pattern recognition, machine learning, and computer vision.

E-mail: wangliang@nlpr.ia.ac.cn

ORCID iD: 0000-0001-5224-8647

\end{document}